\documentclass{article}

 \usepackage[main, final]{neurips_2025}

\usepackage[utf8]{inputenc} 
\usepackage[T1]{fontenc}    

\usepackage{hyperref}       
\usepackage{url}            
\usepackage{booktabs}       
\usepackage{amsfonts}       
\usepackage{nicefrac}       
\usepackage{float}

\usepackage[table, dvipsnames]{xcolor}
\definecolor{lightgreen}{RGB}{220, 245, 220}
\usepackage{enumitem}

\usepackage{times}
\usepackage{latexsym}
\usepackage{array}
\usepackage{multicol}
\usepackage{tablefootnote}
\usepackage{threeparttable}
\usepackage{pifont}
\usepackage{lipsum}
\setlipsum{%
  par-before = \begingroup\color{gray},
  par-after = \endgroup
}

\usepackage{graphicx}
\usepackage{amsmath}
\usepackage{mdframed}
\usepackage{subcaption}

\usepackage{xspace}
\usepackage{kotex} 

\usepackage{listings}  

\lstdefinestyle{jsonbox}{
  basicstyle=\ttfamily\footnotesize,
  columns=fullflexible,
  breaklines=true,
  breakatwhitespace=false,
  keepspaces=true,
  showstringspaces=false,
  tabsize=2,
  upquote=true
}

\usepackage{adjustbox}
\usepackage[breakable]{tcolorbox}

\usepackage{multirow}
\usepackage{svg}
\usepackage{titlesec}

\usepackage{bbm}
\usepackage{bm}

\usepackage{todonotes}    

\usepackage{natbib}

\usepackage{tabularx}
\usepackage{makecell}

\definecolor{customblue}{HTML}{6BE2F6}

\newcommand{\hc}{HyperCLOVA\xspace}
\newcommand{\hcx}{\hc X\xspace}
\newcommand{\hcxtfull}{\hcx 32B Think\xspace}
\newcommand{\hcxtshort}{THINK\xspace}
\newcommand{\hcxofull}{\hcx 8B Omni\xspace}

\newcommand{\eg}{\emph{e}.\emph{g}., }

\title{\hcxtfull}

\def\huggingface{\raisebox{-1.5pt}{\includegraphics[height=1.05em]{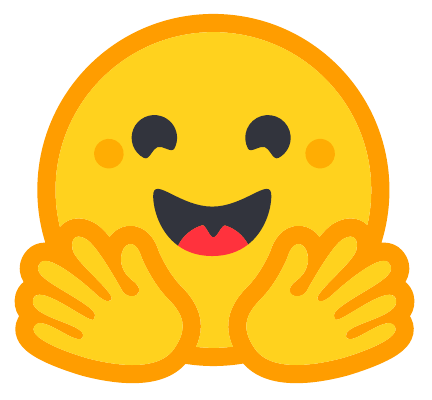}}}

\author{
NAVER Cloud\\
HyperCLOVA X Team\thanks{The complete list of contributors appears in the Contributions and Acknowledgments section. For correspondence, please contact \texttt{dl\_hcx\_technical\_report@navercorp.com}. \textcopyright\ 2025 NAVER Cloud HyperCLOVA X Team. All rights reserved.}\\ 
\small \href{https://huggingface.co/naver-hyperclovax/HyperCLOVAX-SEED-Think-32B}{\huggingface\ \texttt{Huggingface Model Card}}
}

\begin{document}

\maketitle
\begin{abstract}

In this report, we present \hcxtfull, a vision–language model designed with particular emphasis on reasoning within the Korean linguistic and cultural context, as well as agentic ability. \hcxtfull is pre-trained with a strong focus on reasoning capabilities and subsequently post-trained to support multimodal understanding, enhanced reasoning, agentic behaviors, and alignment with human preferences.
Experimental evaluations against comparably sized models demonstrate that our model achieves strong performance on Korean text-to-text and vision-to-text benchmarks, as well as on agent-oriented evaluation tasks. By open-sourcing \hcxtfull, we aim to support broader adoption and facilitate further research and innovation across both academic and industrial communities.

\end{abstract}
\section{Introduction} 

Human reasoning is not solely grounded in textual knowledge but is often informed by visual context. Although large language models (LLMs) have demonstrated solid progress in reasoning ability, their application remains largely confined to scenarios centered on textual information. Expanding training beyond text is indispensable for achieving a deeper and more comprehensive understanding of the real world~\citep{DBLP:conf/icml/HuhC0I24,chen2025janus}. Moreover, as models are more closely integrated into real-world environments, the ability to interact with external tools is becoming essential. Therefore, dedicated training phases aimed at instilling agentic ability are required. In parallel, there is a growing demand for models with a deep understanding of the linguistic and cultural context of Korea, while maintaining the ability to leverage the extensive knowledge available in English. Addressing this requires a training curriculum that incorporates both English and Korean data.

Motivated by these needs, we introduce \hcxtfull (\hcxtshort). \hcxtshort builds on its predecessor~\citep{navercloudhyperclovaxteam2025hyperclovaxthinktechnical} to achieve the capability of understanding both text and vision within a unified framework, while maintaining an emphasis on reasoning in the context of the Korean language and culture, with an added focus on agentic ability. \hcxtshort is a decoder-only Transformer with both text tokens and vision patches projected onto the same continuous embedding space and processed together through standard self-attention. The pre-training curriculum consists of multiple stages with particular attention to honing the ability to reason in Korean. Post-training first instills multimodal capabilities into the model through supervised fine-tuning. Then, the model is further refined in the areas of multimodal reasoning, agentic behavior, and human preference alignment via several reinforcement learning methodologies. This process is conducted in accordance with the NAVER AI Ethics guidelines for enhanced safety.

We evaluate \hcxtshort on various text-to-text (\eg KMMLU, HAERAE-1.0, MMLU, and PIQA), vision-to-text (\eg KoNET, K-DTCBench, SEED-IMG, and DocVQA), and agent (\eg Tau$^2$ and Terminal Bench) benchmarks in Korean and English. We find that \hcxtshort exhibits competitive performance against comparably sized models on text-to-text and vision-to-text benchmarks in Korean, as well as on agent benchmarks. 

\hcxtshort is released as an open-weight model under a custom license that permits commercial use subject to specified conditions. With its competitive performance in Korean and agentic abilities, we present \hcxtshort as a valuable resource for academic and industry partners in both the Korean and global research community.

\section{\hcxtfull (\hcxtshort)}
\begin{figure}[t]
\centering
\includegraphics[width=1.0\textwidth]{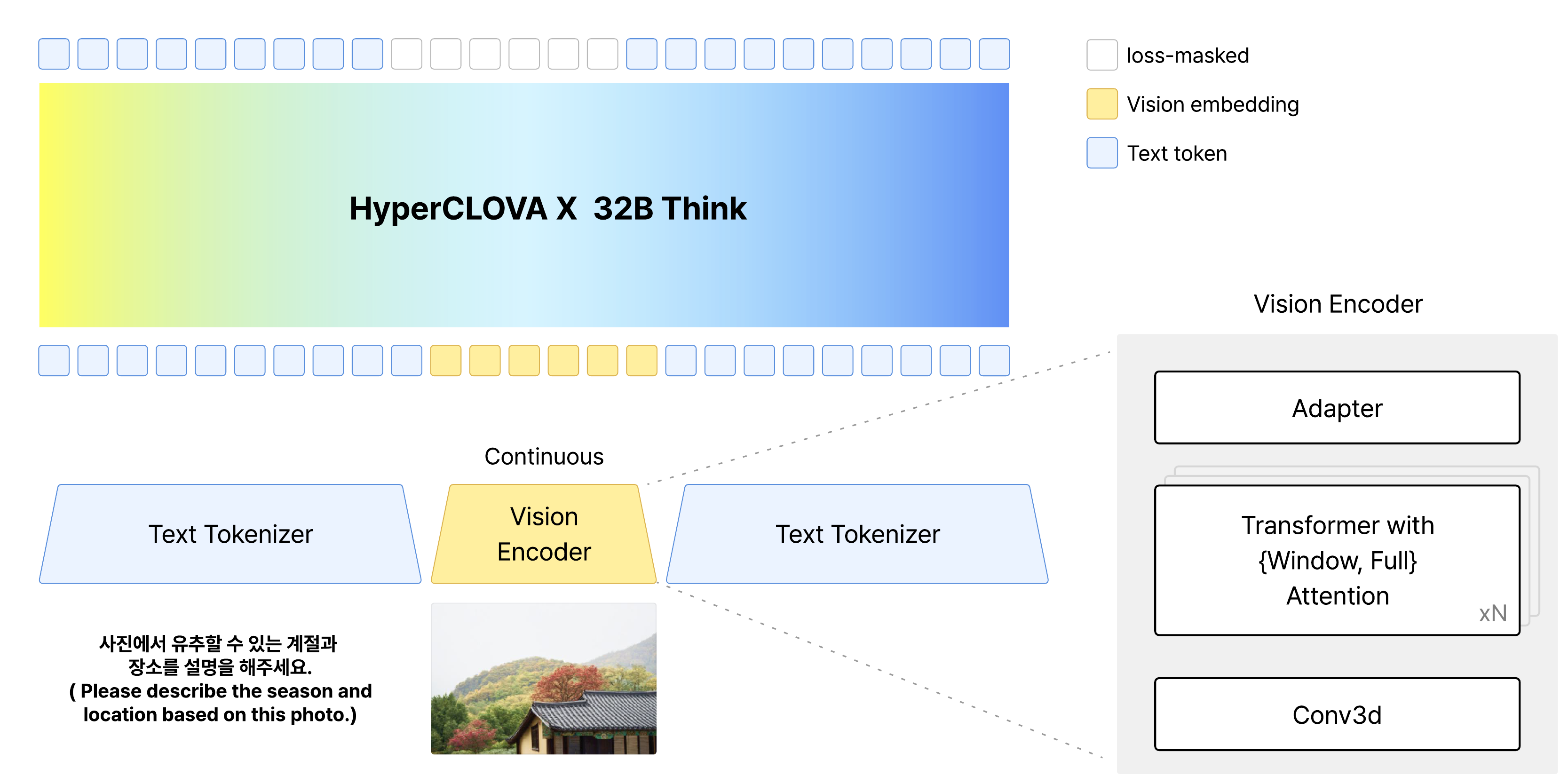}
\caption{Overview of \hcxtfull. Text is tokenized into discrete tokens and mapped to token embeddings, while images are encoded into continuous visual embeddings by the vision encoder. Visual embeddings are projected to the Transformer embedding space and interleaved with text embeddings, enabling joint multimodal processing with a single decoder-only Transformer.}
\label{fig:main_overview}
\end{figure}

As shown in Figure~\ref{fig:main_overview},
\hcxtshort combines a decoder-only Transformer backbone with a text tokenizer and a vision encoder to support unified text-image understanding. In the remainder of this section, we describe the architecture, text tokenizer, and vision encoder in more detail.

\paragraph{Architecture.}
We adopt a standard decoder-only Transformer~\citep{vaswani2017attention} and follow design choices commonly used in recent LLMs to improve its training stability and inference efficiency. Specifically, the model applies RMSNorm~\citep{zhang2019root} as a pre-normalization and adopts SwiGLU ~\citep{shazeer2020gluvariantsimprovetransformer} as the activation function in the feed-forward networks. We use Rotary Positional Embeddings (RoPE)~\citep{su2023roformerenhancedtransformerrotary} for positional encoding with the rotary base parameter set to 500,000 to enhance long-context modeling ability. To improve KV-cache memory efficiency, we employ grouped-query attention~\citep{ainslie2023gqa} with eight query groups. In addition, we remove bias terms from linear layers and untie the input and output embedding weights.
We instantiate the backbone with 72 layers and a hidden size of 5,120, yielding a deep and wide model with sufficient capacity for modeling large-scale knowledge.

For multimodal integration, we employ a streamlined fusion strategy~\citep{liu2023visual, deitke2025molmo, beyer2024paligemma} that minimizes architectural complexity while improving efficiency. Specifically, visual representations are projected into the Transformer embedding dimension and interleaved with text token embeddings. To ensure stability and facilitate seamless collaboration, we maintain the original RoPE configuration of the LLM backbone without introducing intrusive multimodal extensions to the positional encoding. This architecture enables the joint processing of visual and textual representations within a shared representational space while preserving the core efficiency of the pre-trained decoder-only Transformer.

\begin{table}
\centering
\small
\setlength{\tabcolsep}{5pt}

\begin{tabular}{lccccc}
\toprule
\multirow{2.5}{*}{\textbf{Tokenizer}} & \multicolumn{3}{c}{\textbf{English}} & \multicolumn{2}{c}{\textbf{Korean}} \\
\cmidrule(lr){2-4} \cmidrule(lr){5-6}
& \textbf{General} & \textbf{Code} & \textbf{STEM} & \textbf{General} & \textbf{STEM} \\
\midrule
Tiktoken~\citep{tiktoken}  & 4.82 & 3.24 & 3.50 & 1.56 & 1.90 \\
\hcx Think \citep{navercloudhyperclovaxteam2025hyperclovaxthinktechnical} & 4.74 & 3.14 & 3.48 & 1.82 & 2.11 \\
\hcxtfull (\hcxtshort) & 4.55 & 2.72 & 3.35 & 2.15 & 2.18 \\
\bottomrule
\end{tabular}
\caption{Tokenizer compression rate is measured as the average number of characters per subword token across domains, where the general column represents all datasets except for code and math domain datasets. Higher values indicate fewer tokens for the same domain datasets.}
\label{tab:text-tok-compression-rate}
\end{table}

\paragraph{Text-Tokenization.} The tokenization process sequentially employs a pre-tokenizer step and a subword tokenizer step to convert text into a discrete token sequence. The pre-tokenizer step is specifically designed to prevent merging different language scripts, as implemented in DeepSeek-V3~\citep{liu2024deepseek}. Such pre-tokenization rules play a crucial role in determining tokenizer compression efficiency and downstream performance~\citep{dagan-24,Wegmann-25,liu-25,schmidt-25}. An empirical study further indicates that tokenizers yielding a lower upper bound on the Kolmogorov complexity of the tokenized text tend to achieve lower language modeling loss and improved downstream task performance, and that this complexity upper bound is highly sensitive to pre-tokenization rules~\citep{chung-25}. 

Following the approach of LLaMA~\citep{touvron2023llamaopenefficientfoundation} and Qwen~\citep{bai2023qwentechnicalreport}, we implement single-digit tokenization for numbers to prevent performance degradation in code and mathematics tasks~\citep{lee2024digitstodecisions}.
To mitigate \textit{token boundary bias} that may arise from this pre-tokenization~\citep{kudo-2018-subword}, we apply a regularization technique known as StoChasTok~\citep{sims2025stochastok}. We deliberately set the StoChasTok probability to a low value, as high probabilities tend to frequently split words into individual characters, which inflates the token count and consequently reduces training efficiency. We balance this by applying an appropriate probability value.
For subword tokenization, we modify a representative open-source tokenizer, \eg, tiktoken~\citep{tiktoken}, originally optimized for English, Code, and Math. As shown in Table~\ref{tab:text-tok-compression-rate}, these English-centric tokenizers achieve lower compression rates for Korean stems and general text than English. To address this limitation while maintaining performance on the original target domains, we introduce a vocabulary adaptation method comprising two stages: 1) pruning and 2) substitution.

First, the pruning stage removes existing merge rules associated with Korean letters,
which are identified via an automated ISO15924 script detection. Moreover, we remove low-utility rules from non-target languages to maximize the compression rate, resulting in degradation below 1\%. The scope of this trimming is carefully calibrated to ensure that the degradation of compression rate in non-Korean text. Second, the substitution stage inserts new merge rules derived from a trained morpheme-aware subword tokenizer trained on our internal corpus~\citep{navercloudhyperclovaxteam2025hyperclovaxthinktechnical}. Unlike traditional expansion methods~\citep{kim2024efficient} that append new rules at the lowest priority, our substitution stage inserts these rules directly into the priority slots vacated during the pruning stage. As shown in Table~\ref{tab:text-tok-compression-rate}, our tokenizer achieves a substantial increase in Korean compression efficiency (up to 2.15) compared to tiktoken and the tokenizer of the previous version, while maintaining comparable performance across English and STEM domains with minimal trade-offs.

\paragraph{Vision-Encoder.}
For the visual understanding component, \hcxtshort adopts the Vision Transformer (ViT) architecture from Qwen2.5-VL~\citep{bai2025qwen25vltechnicalreport}. The encoder employs a 3D-convolutional layer for patch embedding with a $2 \times 14 \times 14$ kernel, allowing for the unified representation of both static images and video frame sequences~\citep{bai2025qwen25vltechnicalreport}. To maintain computational efficiency when processing high-resolution or long-duration inputs, the encoder incorporates a local-window attention mechanism that constrains interactions to local neighborhoods.

We employ a simplified Vision-Language junction to ensure architectural stability and production efficiency, leveraging the vision encoder that natively supports a multimodal extension of RoPE~\citep{su2024roformer}.
Specifically, we utilize a linear adapter after the vision encoder module to project the encoded dense representations, aligning them with the hidden dimension of the LLM backbone~\citep{liu2023visual, deitke2025molmo, beyer2024paligemma}. This streamlined integration enables joint multimodal processing without requiring intrusive modifications to the core attention mechanism of the decoder-only Transformer.

A primary design objective of \hcxtshort is to improve the efficiency of visual tokens. During training, images are constrained within a $1920 \times 1080$ pixel envelope ($\approx$2.1\,MP), which reduces overall training cost by approximately 53\% in terms of GPU-hours compared to the original Qwen2.5-VL setting ($\approx$12.8\,MP), while maintaining comparable downstream performance. 

While the vision encoder supports a larger pixel budget, we default to the training resolution at inference time to reduce visual token count and improve serving efficiency. For video sequences, we apply a uniform sampling strategy with a maximum of 120 frames, where each frame is constrained to a $378 \times 378$ pixel resolution ($\approx$0.14\,MP). Under this configuration, each video is compressed to a maximum budget of approximately 11\,K tokens, while static images are represented by at most 3\,K tokens.
This approach offers significant computational advantages over contemporary models; for instance, Qwen2.5-VL~\citep{bai2025qwen25vltechnicalreport} and SEED-1.5-VL~\citep{guo2025seed15vltechnicalreport} often allocate over 16K and 82K tokens per input, respectively.

Notably, the vision encoder remains unfrozen throughout training to acquire Korean-centric multimodal capabilities. This full-parameter tuning is essential for internalizing Korean-specific visual contexts, including cultural entities, local landmarks, and high-density Korean-script OCR. By optimizing token allocation and specializing the visual representation for the Korean domain, \hcxtshort achieves a superior balance between multimodal reasoning performance and inference throughput for large-scale deployment.

\section{Pre‑Training} 
A primary focus of the pre-training phase is to endow the model with the capacity to reason in Korean.
For this, we first build a Korean-centered training dataset with a scalable pipeline for cleaning, filtering, masking personal identifiable information (PII), and adding synthetic examples (Section~\ref{subsec:data_preparation}).
We then train with a four-stage curriculum that uses longer contexts and gradually increases the proportion of high-quality reasoning data (Section~\ref{subsec:pretraining_curriculum}).

\subsection{Data Preparation}\label{subsec:data_preparation}
Consistent with the predecessor~\citep{navercloudhyperclovaxteam2025hyperclovaxthinktechnical}, we structured the \hcxtshort data pipeline to prioritize scalability, reusability, and quick refresh. To achieve this, we implement the pipeline with a hybrid processing framework that integrates Datatrove~\citep{penedo2024datatrove} and NeMo-Curator\footnote{\url{https://github.com/NVIDIA-NeMo/Curator}}. This approach minimizes the operational burden of updating datasets while maintaining data integrity. 

Our preprocessing pipeline comprises four stages: (1) collection and normalization stage: we apply lightweight cleaning, render documents into a consistent representation, and map heterogeneous inputs to a unified schema; (2) quality scoring stage: we compute document-level structural and linguistic signals, store them as metadata, and detect and mask PII; (3) filtering stage: we construct stage-specific corpora by combining threshold-based heuristics with learned quality scores; and (4) serialization stage: we serialize the selected documents into sharded files designed for efficient, streaming-based training.

\paragraph{Data Filtering.} Data filtering plays an important role in determining the final training corpus for \hcxtshort in the filtering stage of the data pipeline.
We adapt a two stage pipeline that follows the same structure as the predecessor~\citep{navercloudhyperclovaxteam2025hyperclovaxthinktechnical}. 
The first stage applies lightweight, rule-based heuristics~\citep{redpajama,fineweb-edu} to remove structurally malformed or clearly noisy documents, using document statistics such as symbol ratios, sentence statistics, and boilerplate indicators. This first stage is identical to the previous version.

The second stage performs model-based quality scoring and threshold-based filtering to remove low-quality content that is difficult to identify with heuristic filtering in the previous stage alone. For English data, we retain lightweight scorers based on FastText~\citep{fasttext} and a transformer encoder~\cite{merrick2024arctic}; these models are trained to regress $0$--$5$ quality ratings for efficient large-scale filtering \citep{dclm,fineweb-edu}. This approach has been shown to work well for scoring English educational and STEM documents.

In contrast, our Korean corpus is broad and includes substantial non-STEM, user-facing content such as game guides, consumer reviews, and travel guides. When we applied the English educational scorer to Korean without adaptation, it struggled to distinguish low-quality documents and frequently filtered out useful documents that are not related to the curriculum. To address this mismatch, we train a Korean-specific $0$--$5$ quality regressor on (document, score) pairs obtained from (1) high-quality external corpora (\eg, FineWeb-Edu2~\citep{penedo2025fineweb2pipelinescale}), (2) internal corpora that cover our target domains, and (3) an additional pool of clearly low-quality documents curated via LLM-based annotation. We fine-tune a Qwen3-Embedding-0.6B with a lightweight regression head \citep{qwen3embedding} using our training dataset. Table~\ref{tab:korean_quality_scorer} shows that our Korean scorer consistently improves both accuracy and low-quality recall  over a Korean reproduction of FineWeb-Edu-Classifier. Motivated by recent work~\citep{saada2025data} that the primary gains from data curation come from excluding clearly low-quality documents rather than selecting clear high quality dataset, we choose thresholds for Korean dataset and focus on low-quality removal. Finally, we apply MinHash-based near-duplicate removal after filtering to improve dataset efficiency.
\begin{table*}[t]
\centering
\small
\setlength{\tabcolsep}{2pt}
\begin{tabular}{lcccccc} 
\toprule
\multirow{2}{*}{\textbf{Classifier}} &
\multicolumn{2}{c}{\textbf{FineWeb2 Dataset}} &
\multicolumn{2}{c}{\textbf{In-house Dataset}} &
\multicolumn{2}{c}{\textbf{Average}} \\
\cmidrule(lr){2-3}\cmidrule(lr){4-5}\cmidrule(lr){6-7}
 & Acc. & Recall & Acc. & Recall & Acc. & Recall \\ 
\midrule
\hcx Think \citep{navercloudhyperclovaxteam2025hyperclovaxthinktechnical} 
& 69.0 & 86.8 & 73.5 & 83.2 & 71.3 & 85.0 \\ 

\hcxtfull (\hcxtshort)
& \textbf{86.9} & \textbf{92.5} & \textbf{90.9} & \textbf{92.7} & \textbf{88.9} & \textbf{92.6} \\ 

\bottomrule
\end{tabular}
\caption{Comparison of Korean document quality classifiers measured in accuracy (Acc) and recall (Rec). Labels are annotated $0$--$5$ quality scores and binarized into high-quality ($\ge 3$) and low-quality ($<3$). Recall measures recall on the low-quality class.}
\label{tab:korean_quality_scorer}
\end{table*}

\paragraph{Synthetic Data Generation.} To complement the data filtering strategy and to mitigate the limited availability of high-quality Korean data, we generate synthetic data using two complementary approaches.

First, existing documents are improved and rewritten to enhance their quality. 
    High-quality seed data collected from diverse STEM sources are reformulated to improve coherence while preserving information content. For example, we take high-quality problem--answer pairs and synthesize intermediate reasoning steps, producing self-contained examples that include complete, high-fidelity reasoning traces. For mathematics, we further curate these traces to favor concise reasoning steps, avoiding unnecessarily long solutions.
    
Second, new text is generated from carefully selected high-quality seed data.
    Many high-quality Korean documents contain figures, tables, or diagrams that include essential information. Since removing these elements in the normalization stage makes the remaining text incomplete for understanding the document, we rewrite such documents into self-contained text by using LLMs to generate explicit descriptions from captions and surrounding context. 
    Furthermore, for personally identifiable information, naive masking often degrades fluency and can miss context-dependent PII. We apply a two-step LLM pipeline: (1) detect PII-bearing sentences or spans, then (2) rewrite them with safe, generic substitutes while preserving meaning. This reduces residual PII in the corpus from 89.28\% to 0.13\% without loss of the quality.

Table~\ref{tab:data_composition_stage} illustrates the data composition across training stages, which are detailed in the next subsection. It illustrates how the proportions of general, code, mathematics, and instruction-oriented data are progressively adjusted throughout the training curriculum.

\subsection{Pre-Training Curriculum}\label{subsec:pretraining_curriculum}

\begin{table}[t]
\centering
\begin{tabular}{lcccc}
\toprule
\textbf{Domain} & \textbf{Stage 1} & \textbf{Stage 2} & \textbf{Stage 3} & \textbf{Stage 4} \\
\midrule
General      & 79.4\% & 61.6\%  & 59.1\%  & 17.0\%  \\
Code         & 12.0\% & 20.1\%  & 20.0\%  & 25.2\%  \\
Math         & 8.6\%  & 18.2\%  & 20.0\%  & 25.3\%  \\
Instruction tuning & 0.0\%  & 0.1\%   & 1.0\%   & 32.5\%  \\
\bottomrule
\end{tabular}
\caption{Pre-training data distribution across stages. The General domain dataset comprises all documents excluding code, math, and instruction tuning datasets. The instruction tuning dataset consists of instruction-response pairs. }
\label{tab:data_composition_stage}
\end{table}

We adopt a multi-stage curriculum learning strategy~\citep{olmo-etal-2025-features,navercloudhyperclovaxteam2025hyperclovaxthinktechnical} in which the training signal becomes progressively more demanding across stages. Each stage is designed to emphasize a distinct capability, starting from broad linguistic and factual coverage and then moving toward long-context modeling and reasoning, while gradually increasing both the context window and the proportion of curated high-quality data. Notably, we incorporate reasoning-oriented synthetic data (\eg, coda and math) during pre-training, rather than post-training, which has been observed to improve downstream reasoning performance~\citep{wang2025thinkingaugmentedpretraining}. The detailed data compositions of each training stage are provided in Table~\ref{tab:data_composition_stage}. 

Throughout all stages, we apply Fill in the Middle~\citep{bavarian2022efficient} to $10\%$ of tokens, which strengthens infilling behavior and improves code generation and long-context modeling. Because longer contexts require higher memory and computation costs and lead to different effective batch sizes across stages, we set a different learning rate for each stage to maintain stable and efficient optimization.

\paragraph{Stage 1: Foundation Knowledge Construction.}The first stage builds broad linguistic and factual coverage from a multilingual corpus, primarily Korean and English. Training uses sequences up to 4K tokens and consumes 6 trillion tokens. The learning rate is linearly warmed up from $1.5e^{-5}$ to $3.1e^{-5}$ over the first 2,000 steps, and then kept constant until the end of the first stage.
\paragraph{Stage 2: Context Extension and Quality Up-sampling.}The context length is expanded from 4K to 8K, accompanied by a substantial increase in long-form documents such as academic papers and source code. From this stage onward, high-quality supervised fine-tuning data and chain-of-thought data are mixed into pre-training to establish instruction-following capability. To ensure stable convergence, we apply a cosine decay learning rate schedule and reduce the peak learning rate to approximately 10\% of that used in the previous stage.
\paragraph{Stage 3: Advanced Reasoning and Long-Context Adaptation.} In the third stage, we extend the context length to 32K to further enhance long-context modeling and reasoning abilities. We apply stricter filtering to general Korean and English documents and augment the mixture with reasoning-oriented synthetic data from math and code. To better support very long contexts, we increase the RoPE base frequency from 500K to 5M, and apply cosine decay down to $1.0e^{-5}$.
\paragraph{Stage 4: High-Quality Annealing.}The last stage boosts reasoning ability with a carefully curated high-quality reasoning dataset. We use a carefully selected reasoning-oriented dataset, predominantly composed of math, code, and instruction data. The learning rate starts at $1.0e^{-5}$ and is annealed to one-third of its initial value to maximize reasoning and complex task performance.

\section{Post-Training}
\begin{figure}[t]
\centering
\includegraphics[width=1.0\textwidth]{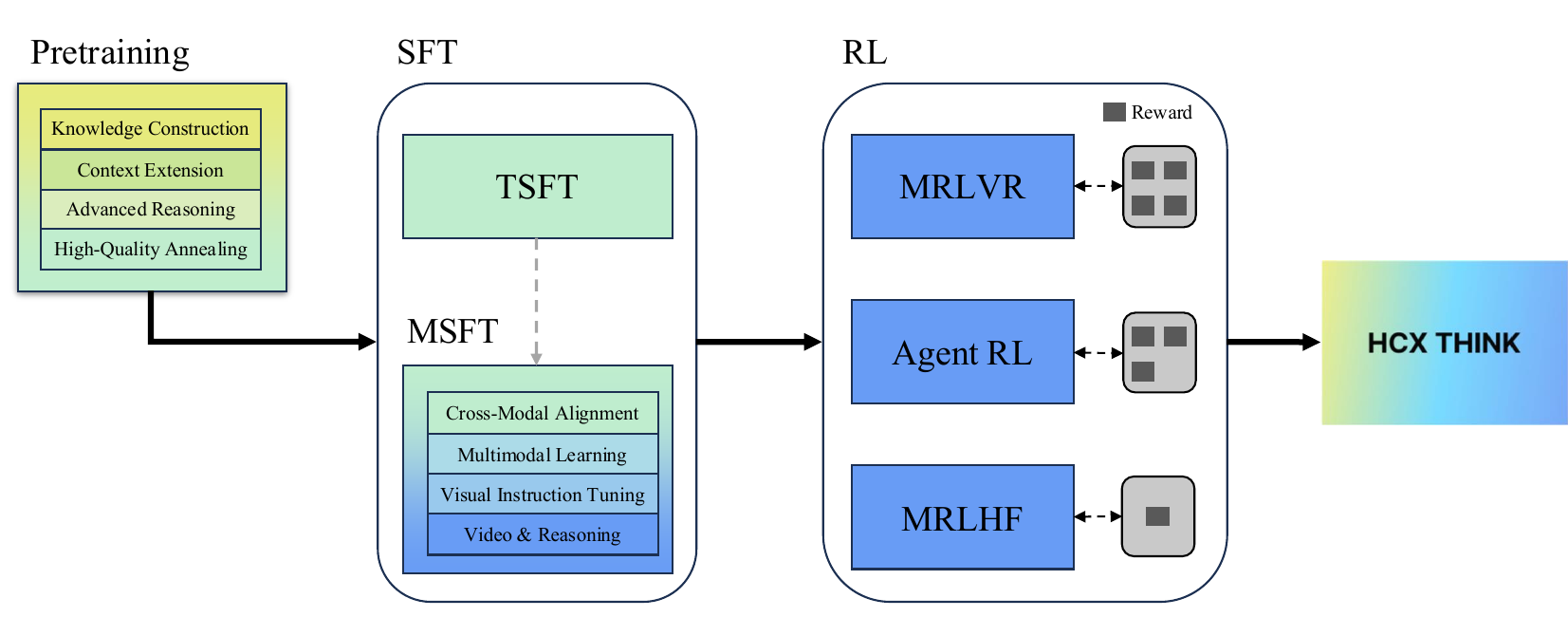}
\caption{Training pipeline of \hcxtfull. 
}
\label{fig:post_pipe}
\end{figure}

As illustrated in Figure~\ref{fig:post_pipe}, the post-training of \hcxtshort is conducted in two phases---supervised fine-tuning (SFT) and reinforcement learning (RL)---with the goal of strengthening reasoning, while also instilling multimodal and agent abilities.
More specifically, we first build a strong foundation for text instruction following and extend it to multimodal capabilities via SFT (Section~\ref{SFT}). We then further refine the model through RL with objectives targeting reasoning, agentic behavior, and alignment with human preferences (Section~\ref{RL}).
This section details the process with particular attention to elements that differ from the predecessor~\citep{navercloudhyperclovaxteam2025hyperclovaxthinktechnical}.

\subsection{Supervised Fine-Tuning (SFT)}
\label{SFT}

\hcxtshort undergoes a two-stage SFT process for enhancing text-based and multimodal instruction following.
Since a well-defined chat template is essential for controlling the generation process, we introduce a unified chat template to be used across various scenarios.
See Appendix \ref{appendidx:chat_templates} for examples.

\paragraph{Text Supervised Fine-Tuning (TSFT).}
TSFT leverages three types of training data: 
\textit{non-reasoning} for general instruction-following;
\textit{reasoning} for multi-step reasoning; and
\textit{agent} for sequential decision making and tool-augmented interactions.
These data types are not strictly disjoint and may overlap in practice. Each data type is processed using a distinct filtering pipeline tailored to its specific characteristics and training objectives.
    To ensure high-quality supervision signals, we strictly retain only those
    trajectories directly sampled using our system that pass all associated test cases.
    As a result, each selected trajectory captures the complete interaction lifecycle,
    including code generation, execution, feedback observation, debugging, and verified
    solution production, while failed or partial runs are excluded.


\paragraph{Multimodal Supervised Fine-Tuning (MSFT).}
MSFT consists of four stages:
\textit{cross-modal alignment} for aligning visual features to the text embedding space;
\textit{multimodal knowledge learning} for broadening visual knowledge and representation capacity; 
\textit{task-oriented instruction tuning} for enhancing task-oriented multimodal interaction; and
\textit{advanced reasoning and video understanding} for establishing long-context multimodal reasoning and temporal dynamics.
This sequential approach is designed to first establish cross-modal alignment and subsequently instill complex reasoning and temporal understanding capabilities. 
To mitigate the degradation of previously acquired text-based reasoning capabilities, TSFT data are continuously incorporated into all stages except Stage 1, with their proportion progressively decreasing from $15.3\%$ in Stage 2 to $6.9\%$ in Stage 4.

\subsection{Reinforcement Learning (RL)}
\label{RL}
\hcxtshort is trained using a combination of reinforcement learning approaches, including multimodal reinforcement learning with verifiable rewards (MRLVR), agent reinforcement learning (Agent-RL), and multimodal reinforcement learning from human feedback (MRLHF).



\paragraph{Multimodal Reinforcement Learning with Verifiable Rewards (MRLVR).}
\hcxtshort develops its core reasoning capabilities through MRLVR across a broad range of tasks and domains. Training is structured into multiple MRLVR stages, each sharing a common verifiable reward framework while targeting a distinct aspect of reasoning behavior: 
\textit{integrated domain RLVR} for general reasoning;
\textit{length-control RLVR} for controlling reasoning length;
\textit{multi-turn RLVR} for checking objective completion; and
\textit{instruction-following RLVR} for instruction-following.

Throughout training, verifiable rewards are complemented by auxiliary reward terms that regulate output format, language consistency, and repetition. To control task difficulty, offline data sampling is applied prior to training to filter out instances that are either trivially easy or excessively challenging, using precomputed pass rates from the initial model. Training is carried out using variants of Group Relative Policy Optimization (GRPO) \citep{shao2024deepseekmath}, augmented with techniques such as dynamic sampling and adaptive clipping \citep{yu2025dapoopensourcellmreinforcement, xi2025bapo} for sample diversity and training stability.

\paragraph{Reinforcement Learning for Agent (Agent-RL).}
We distinguish general agents from SWE agents and train them in different phases. The agents exhibit markedly different rollout characteristics, particularly in rollout duration, interaction depth, and response length. 
On average, SWE agent rollouts are longer (37k vs 4k tokens) with more turns (40 vs 11),
resulting in higher computational cost and longer interaction traces.
Since environment characteristics differ across phases, we design phase-specific environments and reward.
Here, reward is a weighted sum of the following rewards, where the weights are optimized for each domain and task difficulty level:
\textit{environment} for primary evaluation based on rule-based verifiable reward and LLM-as-a-judge-based natural language reward;
\textit{format} for the output format; and
\textit{language consistency} for input-output language matching.

\begin{table}[t]
\centering
\resizebox{0.6\textwidth}{!}{
\begin{tabular}{lcc}
\toprule
 & \textbf{General Agent} & \textbf{SWE Agent} \\
\midrule
Context window size (tokens)                 & 44K   & 128K \\
Max prompt length (tokens)                    & 12K  & 8K \\
Max response length (tokens)                  & 32K  & 120K \\
GRPO group size                      & 8           & 16 \\
\bottomrule
\end{tabular}
}
\caption{Context window size, maximum prompt length, maximum response length, and selected training configurations for agent reinforcement learning. The context window size equals the sum of the prompt and response lengths. Response length includes both model-generated outputs and environment observations.}
\label{tab:agent_length_config}
\vskip -0.1in
\end{table}

Training proceeds in two stages: we first stabilize the core behaviors of a general agent, and then progressively specialize it into an SWE agent.
We perform training using GRPO~\citep{shao2024deepseekmathpushinglimitsmathematical}. To account for the substantial differences between general and SWE agents in interaction length and learning dynamics, we decouple the token length configurations by phase. For general agent training, we use a 44K context window, with the maximum prompt length capped at 12K tokens and the maximum response length capped at 32K tokens. For SWE agent training, we use a 128K context window, with the maximum prompt length set to 8K tokens and the maximum response length set to 120K tokens. Consequently, the total token budget per agent rollout—defined as the sum of prompt tokens and response tokens (including not only model-generated outputs but also environment observation tokens)—differs substantially across phases, as summarized in the Table~\ref{tab:agent_length_config}.

We also adjust GRPO's group size by phase: we use a group size of 8 for general agent training and 16 for SWE agent training. This choice is motivated by the observation that SWE agent tasks exhibit markedly lower success rates and tend to have sparse, high-variance rewards; increasing the group size helps obtain more reliable learning signals and stabilizes policy optimization under these conditions.
From an algorithmic perspective, following DAPO \citep{yu2025dapoopensourcellmreinforcement}, we remove the KL-penalty term that can suppress generation diversity, and we encourage exploration by setting the upper bound of PPO clipping larger than the lower bound. 

Building on these settings, 
Agent-RL adopts the following training strategies, which are designed to reflect the characteristics of agent tasks and to improve training stability, robustness, and ease of deployment in real-world services:
\textit{offline reward filtering} for curating problems of intermediate difficulty for optimal learning;
\textit{thinking mode fusion} for switching between reasoning and non-reasoning modes;
\textit{adaptive rollout sampling parameter} for improving sample efficiency and training stability;
\textit{self-correction via error feedback} for learning recovery behaviors from tool-call exceptions; and
\textit{service caching} for avoiding repeated queries.

\paragraph{Multimodal Reinforcement Learning from Human Feedback (MRLHF).}
We use MRLHF to align the behavior of \hcxtshort with human preferences for improved harmlessness, creativity, and other qualities.
We begin by training a reward model (RM) on a diverse collection of open-source and in-house ranking datasets. In these datasets, response preferences are annotated either by human evaluators or by judge models. Once the reward model is trained, we optimize the policy model to maximize the rewards assigned by the RM using the Proximal Policy Optimization (PPO) \citep{schulman2017proximal} algorithm. Through empirical experiments, we find that PPO, requiring only a single rollout per prompt, enables substantially more optimization steps within a fixed training time compared to variants of GRPO, which rely on multiple rollouts per prompt. Based on this efficiency advantage, we adopt PPO as our primary policy optimization method.

We also introduce an auxiliary reward term to explicitly control both the output format and language of the generated responses. The format-related component enforces adherence to the chat template. The language-related component leverages per-prompt language statistics derived from the chosen responses. If the model generates content in a language that is absent or has a low prior probability under these statistics, a penalty is applied. Both reward model and policy model training are applied exclusively to the final generated responses, while intermediate reasoning steps or think paths are not used for training.

\section{Evaluation}


Recent studies have shown that evaluation of LLMs with reasoning ability are highly sensitive to answer extraction and evaluation protocols~\citep{he2024doespromptformattingimpact, jo2025findinganswersthoughtmatters}. 
Therefore, we test all models in this report using Omni-Evaluator~\footnote{The repository will be publicly released.}, a unified evaluation framework designed to support reproducible benchmarking across various inference engines and modalities.

Omni-Evaluator decouples inference engines from evaluation logic, enabling consistent evaluation across on-device inference (\eg HuggingFace), API-based engines (\eg vLLM, SGLang), and in-house serving systems. In addition, it integrates modality-specific evaluation toolkits into a single framework, which allows unified evaluation across understanding and generation tasks in text, vision, and audio, while ensuring consistency across implementations.

All evaluations are conducted under fixed, version-pinned configurations, including inference engines and evaluation toolkits, ensuring fair and reproducible comparison across models, inference engines, and modalities.

\subsection{Baselines}
We evaluate our model across three benchmark categories: Text-to-Text, Vision-to-Text, and Agent.
For both the Text-to-Text and Vision-to-Text benchmarks, we compare our model against widely adopted open-source vision-language models of comparable scale. 
Specifically, we use Qwen3-VL-32B-Thinking~\citep{yang2025qwen3technicalreport} and InternVL3\_5~38B-Thinking~\citep{wang2025internvl3_5} as baselines, as they are well-established VLMs known for strong multimodal reasoning performance. 
This comparison allows us to assess our model against representative open-source VLMs under both purely textual and vision-conditioned reasoning settings.
To assess our model's fundamental reasoning capabilities beyond the scope of vision language models, we also include EXAONE~4.0~32B~\citep{exaone-4.0} as a text-only baseline.
Although EXAONE is trained as a text-only model without multimodal inputs, it serves as a reference point for comparison between multimodal model and text-only model performance.

Agent benchmarks focus on capabilities beyond standard question answering, including multi-step planning, tool use, and interaction with environments. 
We compare our model not only with open-source baselines but also with substantially larger and more capable commercial models. 
In particular, we include GPT-5.1~\citep{openai_gpt5_system_card_2025} and Qwen3~235B-A22B~\citep{yang2025qwen3technicalreport}, which are widely regarded as strong frontier models for agentic reasoning. 
These models provide a realistic upper-bound on agent performance in practical deployment scenarios.

\subsection{Text-to-Text Benchmarks} 
We aim to develop a model that goes beyond achieving numerical superiority on global benchmarks and instead intrinsically reflects a Korean-centric perspective, incorporating linguistic and cultural inclusiveness. We parse the answer from the generated output using regular expressions. If the answer cannot be found, the sample is considered incorrect.

\definecolor{col}{HTML}{D0EFF9}
\newcommand{\NA}{\cellcolor{blue!3}{--}}
\begin{table}[t]
\centering
\small
\begin{tabular}{ll>{\columncolor{col}}c c c c}
\toprule
\textbf{Language} & \textbf{Dataset} 
& \textbf{\makecell{HyperCLOVA X \\ 32B Think }}
& \textbf{\makecell{EXAONE 4.0 \\ 32B}}
& \textbf{\makecell{Qwen3-VL \\ 32B-Thinking}}
& \textbf{\makecell{InternVL3\_5 \\ 38B-Thinking}} \\
\midrule

\rowcolor{gray!15}
\multicolumn{6}{c}{\textbf{Text-to-Text}} \\
\midrule

\multirow{5}{*}{Korean}
 & KMMLU            & 71.3 & \textbf{75.2} & 47.4(72.1) & 61.2 \\
 & KoBALT           & \textbf{50.6} & 48.3 & 21.1(39.1) & 31.9 \\
 & CLIcK            & \textbf{75.2} & 73.1 & 62.4(72.3) & 62.7 \\
 & HAERAE Bench 1.0 & \textbf{87.4} & 64.3 & 51.5(72.9) & 43.0 \\
 & Flores+ (En$\rightarrow$Ko) & \textbf{31.8} & 30.9 & 24.8 & 26.9 \\
\cmidrule(lr){1-6}

\multirow{4}{*}{English}
 & MMLU      & 87.7 & \textbf{89.9} & 71.5(89.5) & 76.7 \\
 & HellaSwag & 57.2 & \textbf{65.7} & 62.1 & 59.0 \\
 & PIQA      & 76.7 & 75.7 & \textbf{88.0} & 81.1 \\
 & Flores+ (Ko$\rightarrow$En) & 30.9 & 30.2 & 28.2 & \textbf{31.8} \\

\midrule
\rowcolor{gray!15}
\multicolumn{6}{c}{\textbf{Vision-to-Text}} \\
\midrule

\multirow{3}{*}{Korean}
 & KoNET      & \textbf{75.1} & \NA  & 74.3 & 33.9 \\
 & K-MMBench  & 88.1    & \NA & \textbf{91.1} & 81.1 \\
 & K-DTCBench & 93.3    & \NA & \textbf{95.4} & 92.1 \\

\cmidrule(lr){1-6}

\multirow{4}{*}{English}
 & SEED-IMG & 77.9  & \NA & 79.3 & \textbf{88.0} \\
 & LLAVA-W  & 106.4  & \NA & \textbf{111.9} & 100.7 \\
 & TextVQA  & \textbf{85.4}  & \NA & 82.5 & 82.2 \\
 & DocVQA   & 95.5  & \NA & \textbf{96.5} & 93.3 \\

\midrule
\rowcolor{gray!15}
\multicolumn{6}{c}{\textbf{Agent}} \\
\midrule

\multirow{5}{*}{English}
 & Tau$^2$ -- Airline       & \textbf{58.0} & 45.6 & 54.4 & \NA \\
 & Tau$^2$ -- Retail        & \textbf{71.6} & 59.5 & 57.0 & \NA \\       
 & Tau$^2$ -- Telecom       & \textbf{65.1} & 26.1 & 33.3 & \NA \\   
 & Terminal Bench 1.0       & \textbf{21.8} & 10.0 & 11.3 & \NA \\   
 & Terminal Bench Hard      & \textbf{9.9}  & 3.6  & 6.9  & \NA \\   

\bottomrule
\end{tabular}
\caption{Performance comparison across Text-to-Text, Vision-to-Text, and Agent benchmarks (User-simulator: GPT-4.1) in English and Korean.}
\label{tab:benchmark_results}
\end{table}


\begin{table}[t]
\centering
\small
\begin{tabular}{l>{\columncolor{col}}ccc}
\toprule
\textbf{Dataset}
& \textbf{\makecell{HyperCLOVA X \\ 32B Think}}
& \textbf{\makecell{GPT-5.1 \\ (Medium)}}
& \textbf{\makecell{Qwen3 \\ 235B-A22B}}\\
\midrule
\rowcolor{gray!15}
\multicolumn{4}{c}{\textbf{User-simulator: GPT-4.1}} \\
Tau$^2$ -- Airline      &   58.0   &  \textbf{64} & 36.8  \\
Tau$^2$ -- Retail  &  71.6   &  \textbf{78.1} & 51.4  \\       
Tau$^2$ -- Telecom  &  65.1   & \textbf{80.9}  & 20.5  \\   

\midrule
\rowcolor{gray!15}
\multicolumn{4}{c}{\textbf{User-simulator: GPT-5.1} } \\
Tau$^2$ -- Airline      &   58.0   &  \textbf{64} & 41.6  \\
Tau$^2$ -- Retail  &  67.2   &  \textbf{73.6} & 58.4 \\       
Tau$^2$ -- Telecom  &  \textbf{92.3}   & 81.2  & 24.2  \\   

\bottomrule
\end{tabular}
\caption{Tau$^2$ results under different user simulators, comparing \hcxtfull with large-scale frontier models.}
\label{tab:agent_results}
\end{table}

\paragraph{Korean Benchmarks.}
On Korean Text-to-Text benchmarks, \hcxtshort generally outperforms the baselines  across most benchmarks.  
To systematically evaluate Korean text understanding and generation capabilities, we categorize our evaluation suites into three complementary ability dimensions: (1) Korean STEM and professional knowledge, (2) Korean cultural and commonsense knowledge, and (3) Korean linguistic and grammatical competence.

For linguistic and grammatical competence, we use KoBALT~\citep{shin2025kobaltkoreanbenchmarkadvanced}, which probes fine-grained Korean linguistic knowledge (\eg syntax, semantics, pragmatics, phonology, and morphology) and therefore tests structural understanding beyond surface-level recall.
For Korean cultural and commonsense knowledge, we adopt CLIcK~\citep{kim-etal-2024-click} and HAE-RAE Bench (HAERAE-1.0)~\citep{son-etal-2024-hae}, both of which evaluate culturally grounded knowledge, vocabulary, and commonsense reasoning.
Finally, to assess Korean STEM and professional knowledge, we report results on KMMLU~\citep{son-etal-2025-kmmlu}, an expert-level multi-subject benchmark collected from original Korean examinations that measures broad-domain reasoning and professional knowledge expressed in Korean.

Across all Korean benchmarks, \hcxtshort shows a clear advantage over comparable sized open-source vision-language LLMs, such as Qwen3-VL and InternVL. This margin stems from our Korean-aware training recipe, which integrates carefully curated Korean data and training algorithms to enhance performance across all three evaluation dimensions. These results suggest that \hcxtshort has internalized Korean linguistic characteristics as well as the social and cultural context required for faithful expression and reasoning.

\hcxtshort outperforms EXAONE on most Korean text-to-text benchmarks, with the exception of KoBALT. This achievement is particularly notable given that EXAONE is a pure text-only model, which typically benefits from intrinsic advantages in text understanding without the complexity of multimodal objectives. Despite the added complexity of a multimodal architecture, \hcxtshort maintains competitive Korean text reasoning and language understanding.

\paragraph{English Benchmarks.}
In contrast, on English Text-to-Text benchmarks, \hcxtshort shows relatively lower performance than some baseline models.
MMLU~\citep{hendrycks2021measuring} and HellaSwag~\citep{hellaswag} evaluate broad multi-task reasoning and commonsense-based contextual completion in English. On MMLU, \hcxtshort performs competitively with VLM-based models, while showing lower performance than the text-only EXAONE.
On HellaSwag, the \hcxtshort generally shows lower performance than the comparison models, and it also underperforms on physical commonsense reasoning benchmarks such as PIQA~\citep{piqa}. Nonetheless, the performance gap between the \hcxtshort and state-of-the-art models of comparable scale remains moderate, and the model maintains practical levels of reasoning stability and response consistency for real-world use.

\paragraph{Translation Benchmarks.}
We evaluate 1-shot translation performance on the Flores+ benchmark~\citep{nllb-24} using the BLEU metric for the English-Korean pair. For the English-to-Korean direction, we apply Ko-Mecab pre-tokenization to the generated text to compute the scores. As shown in Table~\ref{tab:benchmark_results}, \hcxtshort achieves the best performance in English-to-Korean translation and the second-best performance in Korean-to-English translation. These results demonstrate that \hcxtshort has strong cross-lingual capabilities between Korean and English compared to other baseline models.

\subsection{Vision-to-Text Benchmarks} 

We evaluate Vision-to-Text performance using image-based visual question answering (VQA) benchmarks in both Korean and English. Table~\ref{tab:benchmark_results} summarizes the comparison between \hcxtshort and open-source multimodal LLMs of comparable scale. As with Text-to-Text benchmarks, we also use the same parsing rule.

\paragraph{Image Benchmarks.}

On Korean Vision-to-Text benchmarks, \hcxtshort demonstrates consistent overall performance across a majority of tasks. To systematically assess Korean vision-to-text comprehension, we conducted a comparative analysis using the following benchmarks: (1) the Korean national educational examination~\citep{konet}, (2) general vision-language understanding tasks featuring Korean QA derived from English QA~\citep{kbench}, and (3) Korean document-based VQA~\citep{kbench}.
Specifically, KoNET~\citep{konet} is a comprehensive multimodal benchmark based on Korean national educational standards, encompassing both the Korean General Educational Development (KGED) tests—covering elementary, middle, and high school levels—and the Korean College Scholastic Ability Test (KCSAT). 
K-MMBench~\citep{kbench} evaluates general vision-language understanding capabilities, including object recognition, attribute reasoning, and commonsense inference, while K-DTCBench~\citep{kbench} focuses on document-, table-, and chart-based visual reasoning in Korean.
\hcxtshort  achieves the highest scores among the comparison models on KoNET, indicating strong capabilities in Korean visual understanding and high-difficulty reasoning involving mathematics. On K-MMBench and K-DTCBench, \hcxtshort  ranks second, with a marginal performance gap compared to the top-performing model. Overall, these results demonstrate that \hcxtshort  maintains consistent competitiveness on Korean-centric Vision-to-Text tasks across both education-oriented and general-purpose benchmarks.

We also conduct a comparative analysis on English benchmarks to evaluate diverse visual-language understanding capabilities, including open-ended visual question answering, document understanding, and scene text recognition. We employ four widely adopted benchmarks: SEED-IMG~\citep{seedimg}, LLaVA-W~\citep{llavaw}, TextVQA~\citep{textvqa}, and DocVQA~\citep{docvqa}.
\hcxtshort demonstrates robust text recognition capabilities in both scene-text and dense-document contexts. Notably, \hcxtshort achieves the best performance on TextVQA by a significant margin. However, on the remaining English Vision-to-Text benchmarks, \hcxtshort exhibits relatively lower performance compared to the baseline models.

\paragraph{KCSAT 2026.} To validate the model's real-world reasoning capabilities on the most current academic standards, we evaluate \hcxtshort on the recently conducted KCSAT 2026. We specifically focus on Mathematics, as it presents high-difficulty problems that necessitate the seamless integration of textual instructions with complex visual diagrams. To ensure a robust and stable estimation of the model's performance, we employ consensus@64, deriving the final answer through majority voting across 64 independent generation paths. Furthermore, our evaluation follows a hybrid input strategy: image-based inputs are provided for problems where visual context is vital, while textual representation is also used for others to leverage the full capacity of the backbone. As summarized in Table~\ref{tab:kcsat_2026_math}, \hcxtshort achieves an exceptional score, placing it within the top 4\% of human examinees—a performance level equivalent to the highest tier (1st grade) in the Korean grading system. In particular, \hcxtshort shows strong performance on geometry, where solving problems often requires precise visual–spatial reasoning over 2D and 3D diagrams. These results demonstrate that the model can robustly perform sophisticated, image-based reasoning on novel and challenging academic materials that demand human-level cognitive synthesis.

\begin{table}[t]
\centering
\small
\begin{tabular}{l>{\columncolor{col}}cc}
\toprule
\textbf{Subject (Mathematics)}
& \textbf{\makecell{HyperCLOVA X \\ 32B Think }}
& \textbf{\makecell{1st-Grade Cut-off \\ (Top 4\%)}}\\
\midrule
Probability \& Statistics & 92 & 87 \\
Calculus                  & 89 & 85 \\
Geometry                  & 92 & 85 \\
\bottomrule
\end{tabular}
\caption{Performance of \hcxtfull on the KCSAT 2026 Mathematics exam across elective subjects. The model achieved scores exceeding the estimated 1st-grade (top 4\%) cut-offs in all categories, demonstrating top-level visual reasoning capabilities.}
\label{tab:kcsat_2026_math}
\vskip -0.1in
\end{table}

\paragraph{Video Benchmarks.}
While the primary design philosophy of \hcxtshort focuses on achieving high-fidelity reasoning within the static visual domain, we extend the architecture to support video input processing to ensure service versatility and meet diverse user needs. Video capability is integrated through a uniform sampling strategy of up to 120 frames, capturing temporal dynamics while maintaining computational efficiency. In evaluations on the Video-MME benchmark~\citep{fu2025video} (conducted without subtitles), \hcxtshort achieves a score of 63.4, surpassing established models such as GPT-4V~\citep{openai2023gpt4v} (59.9) and our previous iteration, HCX-Video~\citep{navercloud2025hyperclova} (61.4). Although a performance gap exists relative to models specifically optimized for massive video-reasoning benchmarks (e.g., Qwen3-VL-32B-Thinking at 77.3), \hcxtshort demonstrates superior robustness in localized, real-world content understanding. Notably, in an internal benchmark assessing NAVER TV Content\footnote{\url{https://tv.naver.com/}} comprehension, \hcxtshort scores 67.1, significantly exceeding GPT-4V’s 50.0. These results highlight the model’s practical effectiveness in internalizing the temporal and cultural nuances required for Korean-centric AI services.

\subsection{Agent Benchmarks}
We evaluate agentic capabilities using a collection of agent benchmarks that assess an agent’s ability to perform multi-step reasoning, interact with tools, and maintain task coherence over extended interactions. Specifically, we employ the Tau$^2$ benchmark suite, which consists of \textit{tau$^2$-telecom}, \textit{tau$^2$-retail}, and \textit{tau$^2$-airline}, to evaluate goal completion in interactive multi-turn environments~\cite{tau2bench2025}. These benchmarks are designed to test long-horizon planning, state tracking, and consistent decision-making across sequential interactions.

For SWE-agent–style evaluations that emphasize tool-oriented reasoning in command-line environments, we use TerminalBench~\cite{terminalbench2025}. In this work, TerminalBench 1.0 refers to evaluations conducted using the terminal-bench core implementation version 0.1.1. We additionally report results on TerminalBench Hard, which corresponds to the 47 challenging tasks from the artificial analysis split and is designed to assess robustness to execution errors and recovery from failures.

As shown in Table~\ref{tab:benchmark_results}, \hcxtshort consistently outperforms open-source baseline models across all evaluated agent benchmarks. On the Tau$^2$ tasks, our model achieves higher success rates than both EXAONE~4.0~32B and Qwen3-VL~32B-Thinking, indicating stronger agentic reasoning and more reliable execution of multi-step action sequences. The performance gap is particularly pronounced in scenarios that require longer interaction horizons, suggesting improved state tracking and decision consistency.

Similarly, on TerminalBench and TerminalBench Hard, \hcxtshort demonstrates clear advantages over the baseline models. These results indicate that our model is more effective at integrating reasoning with tool-driven actions even under challenging evaluation settings.

Table~\ref{tab:agent_results} further contextualizes these results by comparing our model with substantially larger commercial and frontier models. While \hcxtshort underperforms relative to GPT-5.1, it consistently outperforms Qwen3~235B-A22B across most agent benchmarks. Given the significant difference in model scale, these results highlight a favorable efficiency–performance trade-off and demonstrate strong agentic capabilities relative to model size.

Detailed evaluation protocols, prompting strategies, and success criteria for each benchmark are provided in Appendix~\ref{appendidx:agent_evaluation_details}.

\subsection{Discussion}
Overall, the evaluation highlights the main strengths of \hcxtshort and its design characteristics.
First, the most prominent strength of \hcxtshort is its strong Korean language understanding and reasoning capability.
Across Korean Text-to-Text benchmarks, \hcxtshort performs on par with or better than the text-only model EXAONE 4.0, despite being a multimodal model. In particular, \hcxtshort shows clear advantages on CLIcK and HAERAE Bench, which require understanding of cultural and social context. These results indicate that \hcxtshort goes beyond a translation-based approach and effectively internalizes linguistic and cultural characteristics of the Korean language.

Second, \hcxtshort also demonstrates competitive performance on Vision-to-Text tasks.
On Korean Vision-to-Text benchmarks, \hcxtshort consistently outperforms the comparison models. This suggests that aligning visual information on top of a text-centric pre-training foundation can be effectively extended to the Korean language setting. Overall, these results support the practical applicability of \hcxtshort as a Korean vision–language reasoning model.

Third, \hcxtshort exhibits strong performance efficiency on agent tasks relative to its model scale.
Across agent benchmarks, \hcxtshort consistently achieves better results than other models of similar size, and in some settings, it shows competitive performance even against large commercial models. This indicates that \hcxtshort has effectively learned agent behaviors that are critical in real-world scenarios, such as long-term state tracking, appropriate tool usage, and recovery from errors.

At the same time, the evaluation also reveals clear limitations of \hcxtshort.
On English-centric Text-to-Text and Vision-to-Text benchmarks, \hcxtshort underperforms compared to some baseline models. This can be understood as a natural trade-off that arises when building a global model while first ensuring strong and stable performance in Korean.

\section{Conclusion}
In this work,  we presented \hcxtfull, a vision-language model trained with an emphasis on reasoning within the context of the Korean language and culture, as well as agentic ability. Experiments against comparably sized models demonstrate that \hcxtfull is competitive on text-to-text (e.g. KoBALT, CLIcK, HAERAE-1.0) and vision-to-text (KoNET and K-MMBench) benchmarks in Korean, as well as agent (Tau$^2$ and Terminal Bench) benchmarks. Given such performance, we expect the public release of \hcxtfull to benefit both the academic and industrial partners.

\hcxtshort is built on a strong text-only backbone that is subsequently extended to vision understanding. This sequential training paradigm enables rapid acquisition of multimodal capabilities by reusing the extensive knowledge learned during text-based pre-training. As a result, competitive vision understanding performance can be achieved with relatively little additional training. However, we observe that as multimodal training progresses, performance on text-based benchmarks gradually degrades, consistent with previous reports on catastrophic forgetting in multimodal large language models~\citep{zhai2023investigating, palme, llavac}. More broadly, this highlights a structural limitation of strictly sequential modality expansion, in which modalities are introduced one after another rather than learned jointly. It becomes increasingly challenging to preserve strong text-based reasoning performance while scaling multimodal capabilities.

In light of this, we do not rely strictly on sequential training in our subsequent model \hcxofull. Instead, we adopt a joint training strategy in which text, vision, and audio modalities are learned together from early stages of training. 
This design mitigates catastrophic forgetting caused by modality extension. 
Please see the \hcxofull report for more details on the omnimodal training strategy.

\bibliographystyle{acl_natbib}
\bibliography{anthology,custom}

\newpage
\appendix

\section*{Contributions and Acknowledgments}
\titlespacing*{\subsection}{0pt}{1ex}{0ex}
\textit{Within each role, \textbf{names are listed in alphabetical order} by first name, followed by the last name.}

\begin{multicols}{2}

\subsection*{Technical Writing}
Cheonbok Park \\
Dongyoon Han \\ 
Geewook Kim \\
Hwiyeol Jo \\
Jaehong Lee \\
Jeonghoon Kim \\
Jin-Hwa Kim \\
Joonsuk Park \\ 
Joosung Lee \\
Sangdoo Yun \\
Sanghyuk Choi \\
Sungwook Jeon \\
Taeho Kil \\ 
Yoonsik Kim \\

\subsection*{Model Research and Training}
Bado Lee \\
Byeongwook Kim  \\
Byungwook Lee \\
Cheonbok Park  \\
Chiheon Ham  \\
Daehee Kim  \\
Donghyeon Ko  \\
DongJu Lee  \\
Geewook Kim  \\
Gichang Lee  \\
Hanbyul Kim  \\
Hangyeol Yu  \\
Hodong Lee  \\
Hyunsoo Ha  \\
Jaehong Lee  \\
Jaemin Han  \\
Jeonghoon Kim  \\
Jinbae Im  \\
Jingu Kang  \\
Jinhyeon Kim  \\
Jisung Wang  \\
Juncheol Kim  \\
Jungwhan Kim  \\
Ka Yeon Song  \\
Kyeongseok Jeong  \\
Min Young Lee  \\
Moonbin Yim  \\
Nako Sung   \\
Ohsung Kwon  \\
Sang Hee Park  \\
Sanghyuk Choi  \\
Seongjin Shin  \\
Shinyoung Joo  \\
Soobee Lee  \\
Sookyo In  \\
Soyoon Kim  \\
Suk Min Seo  \\
Sungju Kim  \\
Taeho Kil  \\
Taehong Min  \\
Taehwan Yoo  \\
Yeontaek Oh  \\
Yoonsik Kim  \\
You Jin Kim  \\
Youngki Hong  \\

\subsection*{Model Evaluation and Analysis}
Chansong Jo \\
Chiheon Ham \\
Daehee Kim \\
Dagyeong Lee \\
DongJu Lee \\
Gayoung Lee  \\
Hagyeong Lee \\
Hangyeol Yu \\
Hwiyeol Jo \\
Hyunhoon Jung \\
Injae Lee \\
Jaemin Han \\
Jeonghyun Lee \\
Jieun Lee \\
Jieun Shin \\
Jingu Kang  \\
Jonghak Kim  \\
Joosung Lee  \\
Jungwhan Kim  \\
Ka Yeon Song  \\
Kiyoon Moon  \\
Min Young Lee  \\
Minchul Song  \\
Minkyoung Kim  \\
Minseong Choi  \\
MinYoung Kim  \\
Munhyong Kim  \\
MyungIn You  \\
Saerim Cho  \\
Shinyoung Joo  \\
Soobee Lee  \\
Suk Min Seo  \\
Taeyong Kim  \\
Yeguk Jin  \\
Yehbin Lee  \\
Yelim Jeong  \\
Yeontaek Oh  \\
Yesol Lee  \\
You Jin Kim  \\
Youngjin Kwon  \\
Youngki Hong  \\

\subsection*{Data}
Chansong Jo \\
Donghyeon Ko \\
Hanbyul Kim \\
Hyunsoo Ha \\
Injae Lee \\
Jieun Lee \\
Jinbae Im \\
Jisung Wang \\
Juncheol Kim \\
Kiyoon Moon \\
Kyeongseok Jeong \\
Minkyoung Kim \\
Minseong Choi \\
Moonbin Yim \\
Munhyong Kim \\
MyungIn You \\
Ohsung Kwon \\
Seongjin Shin \\
Sookyo In \\
Soyoon Kim \\
Sung Ae Lee \\
Sungju Kim \\
Taehong Min \\
Taehwan Yoo \\
Taemin Lim \\
Taeyong Kim \\
Woobin Choi \\
Yeguk Jin \\
Yehbin Lee \\
Youngjun Kim \\
Zoo Hyun Lee \\

\subsection*{Model Serving and Inference}
Ara Jo \\  
Bong-Jin Lee \\ 
Chankyu Lee \\  
Eung Sup Cho \\ 
Gi Nam Kim \\  
Hae Song Cho \\  
Han-Gyu Kim \\  
Hanbae Seo \\  
Hodong Lee \\ 
Hyunjoon Jeong \\  
Jae Ni Jang \\
Jaeeun Kil \\ 
Jaegwang Lee \\ 
Jeongtae Lee \\
Jinhyeon Kim \\  
Joonghoon Kim \\ 
Junga Choi \\ 
Junhee Yoo\\ 
Kwang Sun Jung \\  
Lukas Lee \\  
Minjung Jo \\  
Minsub Kim \\ 
Myungwoo Oh \\  
Ohhyeok Kwon \\  
Sang Hee Park \\  
Se Hee Lee \\  
Seong Chan Hwang \\ 
Seungyeol Lee \\  
Su Jin Lee \\  
Sung Bum Hong \\ 
Sung Woo Lee \\  
Sungjae Lee \\ 
Ui Seon Yu \\ 
Woo Sub Kim \\  
Young Beom Lee \\ 
Young Kyun Jin \\ 
Young Sang Yoo \\  
Young Taek Son  \\
Youngki Kwon \\

\subsection*{Model Planning}
Dagyeong Lee \\
Eunchul Kim \\ 
Gayoung Lee \\ 
Hagyeong Lee \\  
Hyunhoon Jung \\
Jeonghyun Lee \\
Jieun Shin \\
Jonghak Kim \\
JongHyun Lee \\
Matt Yeo \\
Minchul Song \\
MinYoung Kim \\
Saerim Cho \\
Yelim Jeong \\
Yesol Lee \\
Youngjin Kwon \\

\subsection*{Business and Brand Strategy}
Dukmin Jung \\
Kyungmin Lee \\ 
Hyojin Park \\
Sujin Roh \\ 
Misuk Park \\

\subsection*{Residency Program}
Bumkyu Park \\ 
Byung Hyun Lee \\ 
Doohyuk Jang \\
Geeho Kim \\
Hyewon Jeon \\ 
Hyunbin Jin \\ 
Hyungwook Choi \\
Ijun Jang \\
Inju Ha \\
Jewon Yeom \\
Jihwan Kim \\
Jihwan Kwak \\ 
Joonki Min \\ 
Juan Yeo \\ 
Junbeom Kim\\ 
Junyeob Kim\\
Kunhee Kim\\
Kyubyung Chae\\
Kyudan Jung\\
Minha Jhang\\
Sangyoon Lee\\
Sehyun Lee\\
Seunghee Kim\\
Song-ha Jo\\
Suho Ryu\\
Yokyung Lee\\

\subsection*{Internships}

Dong-Jae Lee \\ 
Jihwan Moon \\ 
Jinho Heo \\ 
Jisu Jeon \\ 
Junseo Jang \\ 
Minsik Choi \\
Seulbi Lee \\ 
Singon Kim \\
Sumin Cho \\
Woojin Chung \\
Yunjae Won \\

\end{multicols}


\clearpage
\section*{Appendix}
\section{Chat Templates}
\label{appendidx:chat_templates}

\subsection{Non-Reasoning}
\begin{verbatim}
    <|im_start|>user
    {query}<|im_end|>
    <|im_start|>assistant
    <think>

    </think>

    {response}
    <|im_end|>
\end{verbatim}

\subsection{Reasoning}
\begin{verbatim}
    <|im_start|>user
    {query}<|im_end|>
    <|im_start|>assistant
    <think>
    {reasoning_content}
    </think>

    {response}
    <|im_end|>
\end{verbatim}

\subsection{Agent}
\begin{verbatim}
    <|im_start|>system
    # Tools

    You may call one or more functions to assist with the user 
    query.

    You are provided with function signatures within <tools>
    </tools> XML tags:
    <tools>
    {"type": "function", "function": {"name": "FUNCTION_NAME_1", 
    "description": "DESCRIPTION_1", "parameters": { ... } } }
    {"type": "function", "function": {"name": "FUNCTION_NAME_2", 
    "description": "DESCRIPTION_2", "parameters": { ... } } }
    ...
    </tools>

    For each function call, output the function name and arguments 
    within the following XML format:
    <tool_call>{function-name}
    <arg_key>{arg-key-1}</arg_key>
    <arg_value>{arg-value-1}</arg_value>
    <arg_key>{arg-key-2}</arg_key>
    <arg_value>{arg-value-2}</arg_value>
    ...
    </tool_call><|im_end|>
    <|im_start|>user
    test<|im_end|>
    <|im_start|>assistant
    <think>
    {reasoning_content}
    </think>

    {response}
    <|im_end|>
\end{verbatim}

\subsection{Multimodal}
\begin{verbatim}
    <|im_start|>user
    {multimodal embedding}\n{query}<|im_end|>
    <|im_start|>assistant
    <think>
    {reasoning_content}
    </think>

    {response}
    <|im_end|>
\end{verbatim}
\clearpage
\section{Examples}

\subsection{Vision–Language QA Example}

\begin{figure}[t]
  \centering
  \includegraphics[width=0.6\linewidth]{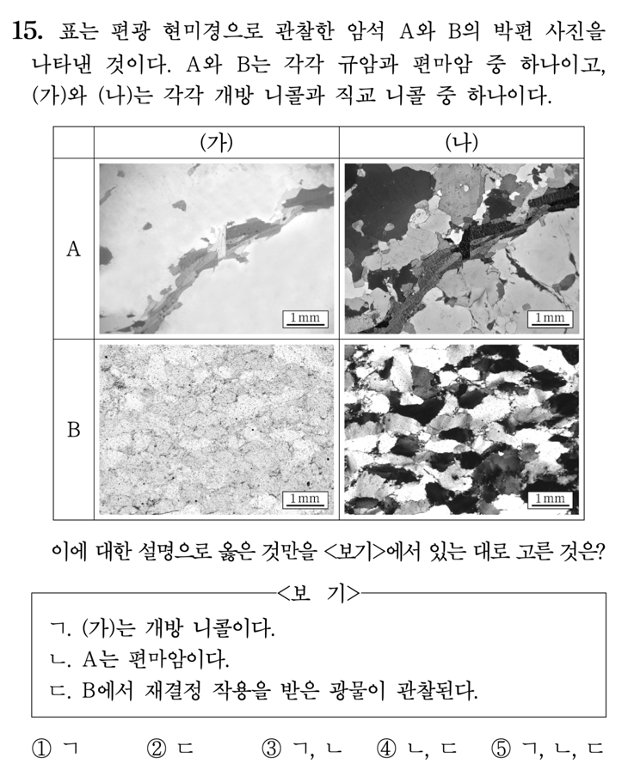}
  \caption{Multimodal Example.}
  \label{fig:vision-figure}
\end{figure}

\paragraph{Model response}
\hspace{1em}
\begin{tcolorbox}[
  breakable,
  colback=white,
  colframe=cyan!70!black,
  colbacktitle=cyan!70!black,
  coltitle=white,
  title=\textbf{HyperCLOVA X THINK (Translated Model Output -- English)},
  fonttitle=\bfseries,
  left=8pt,
  right=8pt,
  top=8pt,
  bottom=8pt,
]

Okay, let’s solve this problem. First, we need to carefully read the question. Thin-section photographs of rocks A and B observed under a polarizing microscope are given, and A and B are each either quartzite or gneiss. In addition, (a) and (b) correspond to either plane-polarized light or cross-polarized light. The task is to choose the correct statements from the options.

First, let’s recall the difference between plane-polarized light and cross-polarized light in a polarizing microscope. Plane-polarized light refers to the state where only the lower polarizer is present, whereas cross-polarized light refers to the state where the upper and lower polarizers are oriented perpendicular to each other. Under cross-polarized light, anisotropic minerals display interference colors, while under plane-polarized light, extinction and general optical properties of minerals can be observed.

Now, looking at the photographs, we need to determine which of (a) and (b) corresponds to plane-polarized light and which to cross-polarized light. Examining the images of A and B, the image of A in (a) appears somewhat dull, with mineral shapes looking indistinct. Under cross-polarized light, minerals usually appear brighter due to interference colors, but (a) looks relatively dark and unclear. In contrast, (b) shows minerals more distinctly, with visible banding and patterns. Since anisotropic minerals exhibit interference colors under cross-polarized light, (b) is more likely to be cross-polarized light. Accordingly, (a) would be plane-polarized light. Therefore, statement ㄱ seems to be correct.

Next, we consider which of A and B corresponds to quartzite or gneiss. Gneiss is a metamorphic rock characterized by gneissic banding, where minerals are arranged in distinct layers or stripes. In the (b) image of A, minerals are clearly arranged in bands, which is characteristic of gneiss. Thus, A is most likely gneiss. On the other hand, B shows a more uniform distribution of minerals in both (a) and (b), without clear banding. Quartzite is composed mainly of quartz and forms through metamorphism accompanied by recrystallization, resulting in uniformly distributed mineral grains. Therefore, B is likely quartzite. Since statement ㄴ claims that A is gneiss, it also appears to be correct.

Finally, statement ㄷ states that minerals that have undergone recrystallization are observed in B. Quartzite forms through recrystallization during metamorphism, so its mineral grains are typically well-developed and uniformly distributed. In the images of B, the minerals appear evenly distributed with clearly developed grains, indicating recrystallization. Therefore, statement ㄷ is also correct.

In summary, statements ㄱ, ㄴ, and ㄷ all appear to be correct. Among the answer choices, option 5 includes all three statements, so the correct answer seems to be 5. Still, it is worth double-checking.

If A is gneiss, the gneissic banding should be visible in the (b) image, and indeed the minerals are arranged in stripes. B is quartzite, showing uniform mineral distribution and evidence of recrystallization, so ㄷ is correct. Since (a) corresponds to plane-polarized light and (b) to cross-polarized light, ㄱ is also correct. Therefore, all statements are correct, and the answer is 5. </think>

\boxed{5}
\end{tcolorbox}

\section{Agent Evaluation}
\label{appendidx:agent_evaluation_details}

This appendix describes the evaluation protocols, benchmark configurations, and execution settings used for agentic capability assessment. We evaluate models using interactive and tool-oriented agent benchmarks, following standardized protocols to ensure reproducibility and fair comparison.

\subsection{Tau\textsuperscript{2} Benchmark Evaluation}

We evaluate interactive agentic reasoning using the Tau$^2$ benchmark suite~\cite{tau2bench2025}, which is designed to assess long-horizon decision-making, state tracking, and goal completion in multi-turn environments. The suite consists of three domains: \textit{tau2-telecom}, \textit{tau2-retail}, and \textit{tau2-airline}, each containing multiple user scenarios with predefined success criteria.

\paragraph{Evaluation Protocol.}
For each task instance, the agent interacts with the environment in a turn-based manner until one of the following conditions is met: the task success condition is satisfied, an explicit failure condition is triggered, or a maximum turn limit is reached. The final outcome is recorded as a binary success or failure signal.

\paragraph{Configurable Evaluation Settings.}
Table~\ref{tab:tau2_eval_settings} summarizes the primary evaluation parameters that can be configured for the Tau$^2$ benchmark. For sampling parameters not listed in Table~\ref{tab:tau2_eval_settings}, the default settings of each model were used. All models were evaluated in reasoning mode.

\begin{table}[h]
\centering
\caption{Evaluation settings for the Tau$^2$ benchmark suite~\cite{tau2bench2025}.}
\label{tab:tau2_eval_settings}
\begin{tabular}{l l p{7cm}}
\toprule
\textbf{Parameter} & \textbf{Value} & \textbf{Description} \\
\midrule
Max turns & 200 & Maximum number of interaction turns allowed per episode \\
Temperature & 0.6 & Sampling temperature used for action generation \\
\bottomrule
\end{tabular}
\end{table}
\newpage
\subsection{TerminalBench Evaluation}

We evaluate tool-oriented agent performance using TerminalBench~\cite{terminalbench2025}, which measures an agent’s ability to solve command-line tasks through iterative tool use and execution feedback. TerminalBench focuses on precise command generation, correct tool usage, and recovery from execution errors.

\paragraph{Benchmarks.}
We report results on TerminalBench 1.0 and TerminalBench Hard. TerminalBench 1.0 corresponds to evaluations conducted using the terminal-bench core implementation version 0.1.1. TerminalBench Hard consists of a subset of 47 challenging tasks selected according to the difficulty definition used by Artificial Analysis, and is intended to stress-test agent robustness in terminal-based execution scenarios\footnote{\url{https://artificialanalysis.ai/methodology/intelligence-benchmarking\#terminal-bench-hard}}.

\paragraph{Evaluation Protocol.}
Each task provides an initial terminal state and a natural language goal description. The agent generates shell commands iteratively, which are executed in a sandboxed environment. After each execution, the agent observes command outputs and error messages. A task is considered successful only if the final system state satisfies all predefined assertions. To ensure fair comparison, all evaluated models utilized the same Terminus-2 agent implementation with reasoning enabled throughout the evaluation.

\subsection{Metrics and Reporting}

For both Tau$^2$ and TerminalBench, the primary evaluation metric is task success rate, defined as the percentage of tasks completed successfully. All results are averaged across task instances within each benchmark unless otherwise specified. Additional diagnostic metrics, such as average interaction length and failure frequency, may be reported to provide further insight into agent behavior.

\end{document}